\documentclass[11pt,a4paper]{article}
\usepackage[hyperref]{acl2019}
\usepackage{times}
\usepackage{latexsym}
\usepackage{amssymb}
\usepackage{amsmath}

\usepackage[pdftex]{graphicx}

\usepackage{url}

\aclfinalcopy 
\def\aclpaperid{48} 

\title{CUED@WMT19:EWC\&LMs}

\author{Felix Stahlberg$^\dag$ \and Danielle Saunders$^\dag$ \and Adri\`a de Gispert$^{\ddagger}$ \and Bill Byrne$^{\ddagger\dag}$\\
 $^\dag$Department of Engineering, University of Cambridge, UK  \\
 $^\ddagger$SDL Research, Cambridge, UK \\
  {\tt \{fs439, ds636, wjb31\}@cam.ac.uk}, {\tt \{agispert, bbyrne\}@sdl.com} \\
  }

\date{}

\begin{document}
\maketitle
\begin{abstract}
  Two techniques provide the fabric of the Cambridge University Engineering Department's (\textbf{CUED}) entry to the \textbf{WMT19} evaluation campaign: elastic weight consolidation (\textbf{EWC}) and different forms of language modelling (\textbf{LMs}). We report substantial gains by fine-tuning very strong baselines on former WMT test sets using a combination of checkpoint averaging and EWC. A sentence-level Transformer LM and a document-level LM based on a modified Transformer architecture yield further gains. As in previous years, we also extract $n$-gram probabilities from SMT lattices which can be seen as a source-conditioned $n$-gram LM.
\end{abstract}

\section{Introduction}

Both fine-tuning and language modelling are techniques widely used for NMT. Fine-tuning is often used to adapt a model to a new domain~\citep{nmt-adaptation-finetuning}, while ensembling neural machine translation (NMT) with neural language models (LMs) is an effective way to leverage monolingual data~\citep{nmt-mono-rnnlm,nmt-mono-rnnlm-csl,simple-fusion}. Our submission to the WMT19 news shared task relies on ideas from these two lines of research, but applies and combines them in novel ways. Our contributions are:

\begin{itemize}
    \item Elastic weight consolidation~\citep[EWC]{nn-ewc} is a domain adaptation technique that aims to avoid degradation in performance on the original domain. We report large gains from fine-tuning our models on former English-German WMT test sets with EWC. We find that combining fine-tuning with checkpoint averaging~\citep{sys-amu-wmt16,nmt-tool-marian} yields further significant gains. Fine-tuning is less effective for German-English.
    \item Inspired by the shallow fusion technique by~\citet{nmt-mono-rnnlm,nmt-mono-rnnlm-csl} we ensemble our neural translation models with neural language models. While this technique is effective for single models, the gains are diminishing under NMT ensembles trained with large amounts of back-translated sentences.
    \item To incorporate document-level context in a light-weight fashion, we propose a modification to the Transformer~\citep{nmt-transformer} that has separate attention layers for inter- and intra-sentential context. We report large perplexity reductions compared to sentence-level LMs under the new architecture. Our document-level LM yields small BLEU gains on top of strong NMT ensembles, and we hope to benefit even more from it in document-level human evaluation.
    \item Even though the performance gap between NMT and traditional statistical machine translation (SMT) is growing rapidly on the task at hand, SMT can still improve very strong NMT ensembles. To combine NMT and SMT we follow \citet{mbr-nmt,ucam-wmt18} and build a specialized $n$-gram LM for each sentence that computes the risk of hypotheses relative to SMT lattices.
    \item While data filtering was central in last year's evaluation~\citep{data-filtering-wmt18,sys-microsoft-wmt18}, in our experiments this year we found that a very simple filtering approach based on a small number of crude heuristics can perform as well as dual conditional cross-entropy filtering~\citep{data-ms-filtering,sys-microsoft-wmt18}.
    \item We confirm the effectiveness of source-side noise for scaling up back-translation as proposed by~\citet{nmt-mono-backtrans-scale}.
\end{itemize}

\section{Document-level Language Modelling}
\label{sec:intra-inter}

MT systems usually translate sentences in isolation. However, there is evidence that humans also take context into account, and judge translations from humans with access to the full document higher than the output of a state-of-the-art sentence-level machine translation system~\citep{nmt-doc-parity}. Common examples of ambiguity which can be resolved with cross-sentence context are pronoun agreement or consistency in lexical choice. This year's WMT competition encouraged submissions of translation systems that are sensitive to cross-sentence context. We explored the use of document-level language models to enhance a sentence-level translation system. We argue that this is a particularly light-weight way of incorporating document-level context. First, the LM can be trained independently on monolingual target language documents, i.e.\ no parallel or source language documents are needed. Second, since our document-level decoder operates on the $n$-best lists from a sentence-level translation system, existing translation infrastructure does not have to be changed -- we just add another (document-level) decoding pass. On a practical note, this means that, by skipping the second decoding pass, our system would work well even for the translation of isolated sentences when no document context is available.


\begin{figure}[t!]
\centering
\small
\includegraphics[scale=0.55]{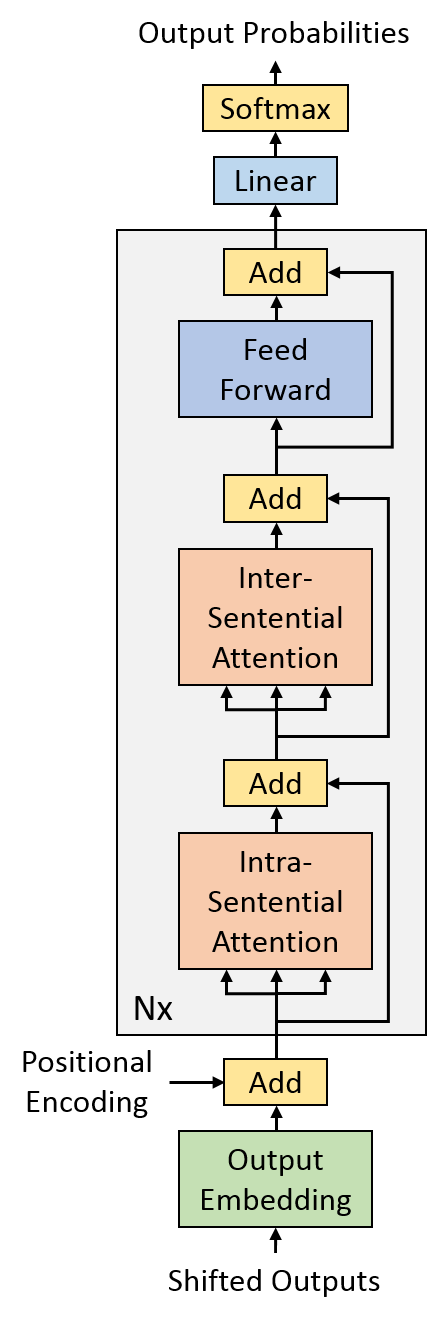}
\caption{Our modified Intra-Inter Transformer architecture with two separate attention layers.}
\label{fig:transformer_lm_intrainter}
\end{figure}

\begin{figure*}[t!]
\centering
\small
\begin{tabular}{l|c@{\hspace{0.4em}}c@{\hspace{0.4em}}c@{\hspace{0.4em}}c@{\hspace{0.4em}}c@{\hspace{0.4em}}c@{\hspace{0.4em}}c@{\hspace{0.4em}}c@{\hspace{0.4em}}c@{\hspace{0.4em}}c@{\hspace{0.4em}}c@{\hspace{0.4em}}c@{\hspace{0.4em}}c@{\hspace{0.4em}}c@{\hspace{0.4em}}c@{\hspace{0.4em}}c}
& Vinyl & destination & : & who & is & actually & buying & records & ? & $</s>$ & Lonely & , & middle-aged & men & love & `???'  \\ \hline
Position & 0 & 1 & 2 & 3 & 4 & 5 & 6 & 7 & 8 & 9 & 10 & 11 & 12 & 13 & 14 & 15 \\
Intra-sentential  & 0 & 0 & 0 & 0 & 0 & 0 & 0 & 0 & 0 & 0 & 1 & 1 & 1 & 1 & 1 & - \\
Inter-sentential & 1 & 1 & 1 & 1 & 1 & 1 & 1 & 1 & 1 & 1 & 0 & 0 & 0 & 0 & 0 & - \\
\end{tabular}
\caption{Intra-sentential and inter-sentential attention masks for an English example from \texttt{news-test2017}. Document-level context helps to predict the next word (`vinyl').}
\label{fig:attention-masks}
\end{figure*}

Our document-level LMs are trained on the concatenations of all sentences in target language documents, separated by special sentence boundary tokens. Training a standard Transformer LM~\citep{nmt-transformer} on this data already yields significant reductions in perplexity compared to sentence-level LMs. However, the attention layers have to capture two kinds of dependencies 
-- the long-range cross-sentence context and the short-range context within the sentence. Our modified Intra-Inter Transformer architecture (Fig.~\ref{fig:transformer_lm_intrainter}) splits these two responsibilities into two separate layers using masking. The ``Intra-Sentential Attention'' layer only allows to attend to the previous tokens in the current sentence, i.e.\ the intra-sentential attention mask activates the tokens between the most recent sentence boundary marker and the current symbol. The ``Inter-Sentential Attention'' layer is restricted to the tokens in all previous {\em complete} sentences, i.e.\ the mask enables all tokens from the document beginning to the most recent sentence boundary marker. As usual~\citep{nmt-transformer}, during training the attention masks are also designed to prevent attending to future tokens. Fig.~\ref{fig:attention-masks} shows an example of the different masks. Note that as illustrated in Fig.~\ref{fig:transformer_lm_intrainter}, both attention layers are part of the same layer stack which allows a tight integration of both types of context. An implication of this design is that they also use the same positional embedding -- the positional encoding for the first unmasked item for intra-sentential attention may not be zero. For example, `Lonely' has the position 10 in Fig.~\ref{fig:attention-masks} although it is the first word in the current sentence.

We use our document-level LMs to rerank $n$-best lists from a sentence-level translation system.
Our initial document is the first-best sentence hypotheses. We greedily replace individual sentences with lower-ranked hypotheses (according to the translation score) to drive up 
a combination of translation and document LM scores. We start with the sentence with the minimum difference between the first- and second-best translation scores. We stop when the translation score difference to the first-best translation exceeds a threshold.\footnote{Tensor2Tensor implementation: \url{https://github.com/fstahlberg/ucam-scripts/blob/master/t2t/t2t_refine_with_glue_lm.py}}

\section{Experimental Setup}

Our experimental setup is essentially the same as last year~\citep{ucam-wmt18}: Our pre-processing includes Moses tokenization, punctuation normalization, truecasing, and joint subword segmentation using byte pair encoding~\citep{nmt-bpe} with 32K merge operations. We compute cased BLEU scores with \texttt{mteval-v13a.pl} that are directly comparable with the official WMT scores.\footnote{\url{http://matrix.statmt.org/}} Our models are trained with the TensorFlow~\citep{nn-tensorflow} based Tensor2Tensor~\citep{nmt-tool-t2t} library and decoded with our SGNMT framework~\citep{sgnmt1,sgnmt2}. We delay SGD updates~\citep{danielle-syntax} to use larger training batch sizes than our technical infrastructure\footnote{The Cambridge HPC service (\url{http://www.hpc.cam.ac.uk/}) allows parallel training on up to four physical P100 GPUs.} would normally allow with vanilla SGD by using the \texttt{MultistepAdam} optimizer in Tensor2Tensor. We use Transformer~\citep{nmt-transformer} models in two configurations (Tab.~\ref{tab:transformer-setups}). Preliminary experiments are carried out with the `Base' configuration while we use the `Big' models for our final system. We use \texttt{news-test2017} as development set to tune model weights and select checkpoints and \texttt{news-test2018} as test set.

\subsection{ParaCrawl Corpus Filtering}

\citet{data-ms-filtering,sys-microsoft-wmt18} reported large gains from filtering the ParaCrawl corpus. This year, the WMT organizers made version 3 of the ParaCrawl corpus available. We compared two different filtering approaches on the new data set. First, we implemented dual cross-entropy filtering~\citep{data-ms-filtering,sys-microsoft-wmt18}, a sophisticated data selection criterion based on neural language model and neural machine translation model scores in both translation directions. In addition, we used the ``naive'' filtering heuristics proposed by~\citet{ucam-wmt18}:
\begin{itemize}
\item Language detection~\citep{langdetect} in both source and target language.
\item No words contain more than 40 characters.
\item Sentences must not contain HTML tags.
\item The minimum sentence length is 4 words.
\item The character ratio between source and target must not exceed 1:3 or 3:1.
\item Source and target sentences must be equal after stripping out non-numerical characters.
\item Sentences must end with punctuation marks.
\end{itemize}
Tab.~\ref{tab:filtering} indicates that our systems benefit from ParaCrawl even without filtering (rows 1 vs.\ 2). Our best `Base' model uses both dual and naive filtering. However, the difference between filtering techniques diminishes under stronger `Big' models with back-translation (rows 6 and 7).

\begin{table}
\centering
\small
\begin{tabular}{|l|l|l|}\hline
 & \textbf{Base} & \textbf{Big} \\ \hline
T2T HParams set & \texttt{trans.\_base} & \texttt{trans.\_big} \\
\# physical GPUs & 4 & 4 \\
Batch size & 4,192 & 2,048 \\
SGD delay factor & 2 & 4 \\
\# training iterations & 300K & 1M \\
Beam size & 4 & 8 \\
    \hline
\end{tabular}
\caption{Transformer setups.}\label{tab:transformer-setups}
\end{table}

\begin{table*}
\centering
\small
\begin{tabular}{r|l|l|c|cccc|}\cline{2-8}
& \textbf{Model} & \textbf{ParaCrawl} & \textbf{Naive} &  \multicolumn{4}{c|}{\textbf{BLEU}} \\
     & &  & \textbf{filtering} &  \textbf{test15} & \textbf{test16} & \textbf{test17} & \textbf{test18} \\ \cline{2-8}
\footnotesize{1} & Base & No & & 29.3 & 34.1 & 27.8 & 41.9 \\ \cline{2-8}
\footnotesize{2} & Base & Full & & 30.0 &  35.3 & 28.2 & 43.1 \\
\footnotesize{3} & Base & Full & \checkmark & 30.3 & 35.6 & 28.6 & 43.5 \\
    \cline{2-8}
\footnotesize{4} & Base & Dual x-ent filtering & & 30.2 & 35.5 & 28.7 & 43.6 \\
\footnotesize{5} & Base & Dual x-ent filtering & \checkmark & 30.6 & 35.7 & 28.8 & 43.8 \\
    \cline{2-8}
\footnotesize{6} & Big (with back-translation) & Full & \checkmark & 32.4 & 38.5 & 31.2 & 46.6 \\
\footnotesize{7} & Big (with back-translation) & Dual x-ent filtering & \checkmark & 32.7 & 38.1 & 31.1 & 46.6 \\
\cline{2-8}
\end{tabular}
\caption{Comparison of ParaCrawl filtering techniques. The rest of the training data is over-sampled to roughly match the size of the filtered ParaCrawl corpus. In the `Dual x-ent filtering' experiments we selected the 15M best sentences according the dual cross-entropy filtering criterion of \citet{data-ms-filtering}.}\label{tab:filtering}
\end{table*}

\section{Results}

\begin{table*}
\centering
\small
\begin{tabular}{r|ccc|c|cccc|}\cline{2-9}
& \textbf{news-2016} & \textbf{news-2017} & \textbf{news-2018} & \textbf{Noise}    & \multicolumn{4}{c|}{\textbf{BLEU}} \\
     & \textbf{(35M sentences)} & \textbf{(20M sentences)} & \textbf{(37M sentences)} & &  \textbf{test15} & \textbf{test16} & \textbf{test17} & \textbf{test18} \\ \cline{2-9}
\footnotesize{1} & &  & & & 30.2 & 35.7 & 28.7 & 43.8 \\ \cline{2-9}
\footnotesize{2} &    & \checkmark & & & 30.8 & 36.2 & 29.8 & 44.3 \\
\footnotesize{3} &    & & \checkmark  & & 30.4 & 35.8 & 29.4 & 43.2 \\
\footnotesize{4} &    & \checkmark & \checkmark & & 30.3 & 35.9 & 29.5 & 43.1 \\ \cline{2-9}
\footnotesize{5} &    & \checkmark & & \checkmark & 31.0 & 36.6 & 29.7 & 44.8 \\
\footnotesize{6} &    & & \checkmark  & \checkmark & 30.7 & 36.6 & 29.5 & 44.7 \\
\footnotesize{7} &     & \checkmark & \checkmark & \checkmark & 30.6 & 36.6 & 29.5 & 44.4 \\
\cline{2-9}
\footnotesize{8}     & \checkmark & \checkmark &  & \checkmark & 31.3 & 37.4 & 30.0 & 45.2 \\
\footnotesize{9}     & \checkmark & \checkmark & \checkmark & \checkmark & 31.3 & 37.3 & 30.3 & 45.2 \\
    \cline{2-9}
\end{tabular}
\caption{Using different corpora for back-translation. We back-translated with a `base' model for \texttt{news-2017} and the big single Transformer model of \citet{ucam-wmt18} for \texttt{news-2016} and \texttt{news-2018}.}\label{tab:backtrans}
\end{table*}

\subsection{Back-translation}

Back-translation~\citep{backtranslation} is a well-established technique to use monolingual target language data for NMT. The idea is to automatically generate translations into the source language with an inverse translation model, and add these synthetic sentence pairs to the training data. A major limitation of vanilla back-translation is that the amount of synthetic data has to be balanced with the amount of real parallel data~\citep{backtranslation,sys-uedin-wmt16,nmt-mono-backtrans-ana}. \citet{nmt-mono-backtrans-scale} had overcome this limitation by adding random noise to the synthetic source sentences. Tab.~\ref{tab:backtrans} shows that using noise improves the BLEU score by between 0.5 and 1.5 points on the \texttt{news-test2018} test set (rows 2-4 vs.\ 5-7).\footnote{We use Sergey Edunov's \texttt{addnoise.py} script available at \url{https://gist.github.com/edunov/d67d09a38e75409b8408ed86489645dd}} Our final model uses a very large number (92M) of (noisy) synthetic sentences (row 9), although the same performance could already be reached with fewer sentences (row 8).

\subsection{Fine-tuning with EWC and Checkpoint Averaging}

\begin{table}
\centering
\small
\begin{tabular}{r|l|c|c|c|}\cline{2-5}
& \textbf{Fine-tuning} & \textbf{Checkpoint} & \multicolumn{2}{|c|}{\textbf{BLEU (test18)}} \\
& & \textbf{averaging} & \textbf{En-De} & \textbf{De-En}  \\ \cline{2-5}
\footnotesize{1} & No & & 46.7 & 46.5 \\
\footnotesize{2} & No & \checkmark & 46.6 & 46.4  \\
    \cline{2-5}
\footnotesize{3} & Cont'd train. & & 47.1 & 46.6 \\
\footnotesize{4} & Cont'd train. & \checkmark & 47.3 & 46.8 \\
    \cline{2-5}
\footnotesize{5} & EWC & & 47.1 & 46.4 \\
\footnotesize{6} & EWC & \checkmark & 47.8 & 46.8 \\
    \cline{2-5}
\end{tabular}
\caption{Fine-tuning our models on former WMT test sets using continued training and EWC.}\label{tab:finetuning}
\end{table}

\begin{table*}
\centering
\small
\begin{tabular}{|l|l|cccc|cccc|}\hline
 &   & \multicolumn{8}{c|}{\textbf{Perplexity (per subword)}} \\
\multicolumn{1}{|c|}{\textbf{Model}} & \multicolumn{1}{c|}{\textbf{Context}}  & \multicolumn{4}{c|}{\textbf{German}}  & \multicolumn{4}{c|}{\textbf{English}} \\
     & &   \textbf{test15} & \textbf{test16} & \textbf{test17} & \textbf{test18} &   \textbf{test15} & \textbf{test16} & \textbf{test17} & \textbf{test18} \\ \hline
    Standard (Big) & Sentence-level & 36.23 & 35.69 & 36.17 & 34.77 & 39.94 & 37.19 & 35.34 & 42.38  \\
    Standard(Big)  & Document-level & 26.63 & 27.85 & 25.43 & 28.36 & 43.37 & 34.55 & 31.27 & 39.74  \\
    Intra-Inter (Big) & Document-level & 23.54 & 22.39 & 22.05 & 22.56 & 34.25 & 31.16 & 29.31 & 34.47 \\
    \hline
\end{tabular}
\caption{Language model perplexities of different neural language models. `Intra-Inter' denotes our modified Transformer architecture from Sec.~\ref{sec:intra-inter}. The standard model has 448M parameters, Intra-Inter has 549M parameters.}\label{tab:document-ppl}
\end{table*}

\begin{table*}
\centering
\small
\begin{tabular}{r|l|c|cc||c|cc|}\cline{2-8}
& & \multicolumn{3}{c||}{\textbf{English-German}} & \multicolumn{3}{c|}{\textbf{German-English}} \\
& & \textbf{Base} & \multicolumn{2}{c||}{\textbf{Big (with EWC)}} & \textbf{Base} & \multicolumn{2}{c|}{\textbf{Big (with EWC)}} \\
&  & \textbf{Single} & \textbf{Single} & \textbf{4-Ensemble} & \textbf{Single} & \textbf{Single} & \textbf{4-Ensemble}  \\ \cline{2-8}
\footnotesize{1} & Using back-translation? & No & Yes & Yes & No & Yes & Yes \\
\cline{2-8}
\footnotesize{2} & NMT & 43.8 & 47.8 & 48.8 & 40.7 & 47.4 & 48.3  \\
\footnotesize{3} & + Sentence-level LM & 44.7 & 47.8 & 48.8 & 41.4 & 47.6 & 48.3 \\
\footnotesize{4} & \hspace{0.7em}+ PBSMT (MBR-based) & 45.1 & 48.0 & 49.1 & 42.1 & 47.6 & 48.5 \\
\footnotesize{5} & \hspace{1.4em}+ Document-level Intra-Inter LM & 45.7 & 47.6 & 49.3 & 42.1 & 47.3 & 48.6 \\
\cline{2-8}
\end{tabular}
\caption{Using different kinds of language models for translation on \texttt{news-test2018}. The PBSMT baseline gets 26.7 BLEU on English-German and 27.5 BLEU on German-English.}\label{tab:nmt-lm}
\end{table*}

Fine-tuning~\citep{nmt-adaptation-finetuning} is a domain adaptation technique that first trains a model until it converges on a training corpus A, and then continues training on a usually much smaller corpus B which is close to the target domain. Similarly to \citet{sys-rwth-wmt18,sys-jhu-wmt18}, we fine-tune our models on former WMT test sets (2008-2016) to adapt them to the target domain of high-quality news translations. Due to the very small size of corpus B, much care has to be taken to avoid over-fitting. We experimented with different techniques that keep the model parameters in the fine-tuning phase close to the original ones. First, we fine-tuned our models for about 1K-2K iterations (depending on the performance on the \texttt{news-test2017} dev set) and dumped checkpoints every 500 steps. Averaging all fine-tuning checkpoints together with the last unadapted checkpoint yields minor gains over fine-tuning without averaging (rows 3 vs.\ 4 in Tab.~\ref{tab:finetuning}). However, we obtain the best results by combining checkpoint averaging with another regularizer -- elastic weight consolidation~\citep[EWC]{nn-ewc} -- that explicitly penalizes the distance of the model parameters $\theta$ to the optimized but unadapted model parameters $\theta^*_{A}$. The regularized training objective according EWC is:
\begin{equation}
L(\theta) = L_B(\theta) + \lambda \sum_i F_i (\theta_i - \theta^*_{A,i})^2
\label{eq:ewc}
\end{equation}
where $L_B(\theta)$ is the normal cross-entropy training loss on task B and $F_i=\mathbb{E} \big[ \nabla^2  L_A(\theta_i)\big] $ is an estimate of task $A$ Fisher information, which represents the importance of parameter $\theta_i$ to $A$. On English-German, fine-tuning with EWC and checkpoint averaging yields an 1.1 BLEU improvement (rows 1 vs.\ 6 in Tab.~\ref{tab:finetuning}). Gains are generally smaller on German-English.

\subsection{Language modelling}

We introduced our new Intra-Inter Transformer architecture for document-level language modelling in Sec.~\ref{sec:intra-inter}. Tab.~\ref{tab:document-ppl} shows that our architecture achieves much better perplexity than both a sentence-level language model and a document-level  vanilla Transformer model. Tab.~\ref{tab:nmt-lm} summarizes our translation results with various kinds of language models. Adding a Transformer sentence-level LM to NMT helps for the single Base model without back-translation, but is less effective on top of (ensembles of) Big models with back-translation (row 2 vs.\ 3). Extracting $n$-gram probabilities from traditional PBSMT lattices as described by~\citet{mbr-nmt} and using them as source-conditioned $n$-gram LMs yields gains even on top of our ensembles (row 4). Our document-level Intra-Inter language models improve the ensembles and the single En-De Base model, but hurt performance slightly for the single Big models (row 5).

\section{Related Work}

\paragraph{Regularized fine-tuning} Our approach to fine-tuning is a combination of EWC~\citep{nn-ewc} and checkpoint averaging~\citep{sys-amu-wmt16,nmt-tool-marian}. In our context, both methods aim to avoid {\em catastrophic forgetting}\footnote{Catastrophic forgetting occurs when the performance on the specific domain is improved after fine-tuning, but the performance of the model on the general domain has decreased drastically.}~\citep{nmt-adaptation-cf-investigation,nmt-adaptation-cf} and over-fitting by keeping the adapted model close to the original, and can thus be seen as {\em regularized} fine-tuning techniques. \citet{nmt-adaptation-secretly-kd,nmt-adaptation-kd} regularized the output distributions during fine-tuning using techniques inspired by knowledge distillation~\citep{kd-original,kd-hinton,kd-nmt}. \citet{nmt-adaptation-l2} applied standard L2 regularization\index{L2 regularization} and a variant of dropout to domain adaptation. EWC as generalization of L2 regularization has been used for NMT domain adaptation by~\citet{nmt-adaptation-ewc,nmt-adaptation-ewc-danielle}. In particular, \citet{nmt-adaptation-ewc-danielle} showed that EWC is not only more effective than L2 in reducing catastrophic forgetting but even yields gains on the general domain when used for fine-tuning on a related domain. 

\paragraph{Document-level MT} Various techniques have been proposed to provide the translation system with inter-sentential context, for example by initializing encoder or decoder states~\citep{nmt-doc-init}, using multi-source encoders~\citep{nmt-doc-eval,nmt-doc-multisrc-src}, as additional decoder input~\citep{nmt-doc-init}, with memory-augmented neural networks~\citep{nmt-doc-cache,nmt-doc-mem,nmt-doc-cache-trg}, hierarchical attention~\citep{nmt-doc-han,nmt-doc-selective-att}, deliberation networks~\citep{nmt-doc-delib}, or by simply concatenating multiple source and/or target sentences~\citep{nmt-doc-concat,nmt-doc-eval}. Context-aware extensions to Transformer encoders have been proposed by~\citet{nmt-doc-transformer-enc1,nmt-doc-transformer-enc2}. Techniques also differ in whether they use source context only~\citep{nmt-doc-multisrc-src,nmt-doc-init,nmt-doc-transformer-enc1,nmt-doc-transformer-enc2}, target context only~\citep{nmt-doc-cache,nmt-doc-cache-trg}, or both~\citep{nmt-doc-eval,nmt-doc-mem,nmt-doc-han,nmt-doc-concat,nmt-doc-selective-att}. Several studies on document-level NMT indicate that automatic and human sentence-level evaluation metrics often do not correlate well with improvements in discourse level phenomena~\citep{nmt-doc-eval,nmt-doc-parity,nmt-doc-test-set}. Our document-level LM approach is similar to the work of \citet{nmt-doc-delib} in that cross-sentence context is only used in a second pass to improve translations from a sentence-level MT system. Our method is light-weight as, similarly to \citet{nmt-doc-concat}, we do not modify the architecture of the core NMT system.

\paragraph{NMT-SMT hybrid systems} Popular examples of combining a fully trained SMT system with independently trained NMT are rescoring and reranking methods~\citep{hybrid-nbest,sgnmt,nmt-adaptation-sgnmt,gec-jd-hybrid,hybrid-reranking-german,hybrid-smorgasbord,hybrid-pb-forced}, although these models may be too constraining if the neural system is much stronger than the SMT system. Loose combination schemes include the edit-distance-based system of \citet{edit-dist-wmt16} or the minimum Bayes-risk approach of \citet{mbr-nmt} we adopted in this work. NMT and SMT can also be combined in a cascade, with SMT providing the input to a post-processing NMT system~\citep{hybrid-pre-translation,hybrid-nmt-multisource} or vice versa~\citep{hybrid-neural-pre-translation}.  \citet{hybrid-smt-advise,hybrid-smt-advise2} interpolated NMT posteriors with word recommendations from SMT and jointly trained NMT together with a gating function which assigns the weight between SMT and NMT scores dynamically. The AMU-UEDIN submission to WMT16 let SMT take the lead and used NMT as a feature in phrase-based MT~\citep{sys-amu-wmt16}. In contrast, \citet{hybrid-technical-terms} translated most of the sentence with an NMT system, and just used SMT to translate technical terms in a post-processing step. \citet{hybrid-nmt-smt-search} proposed a hybrid search algorithm in which the neural decoder expands hypotheses with phrases from an SMT system.

\section{Conclusion}

\begin{table}
\centering
\small
\begin{tabular}{|l|c|}\hline
\multicolumn{2}{|c|}{\textbf{English-German}} \\
\textbf{Team} & \textbf{BLEU} \\ \hline
MSRA & 44.9 \\
Microsoft & 43.9 \\
NEU & 43.5 \\
\textbf{UCAM} & \textbf{43.0} \\
Facebook FAIR &  42.7 \\
JHU & 42.5 \\
eTranslation & 41.9 \\
\multicolumn{2}{|c|}{{\em 8 more...}} \\
    \hline
\end{tabular}\hspace{0.5em}
\begin{tabular}{|l|c|}\hline
\multicolumn{2}{|c|}{\textbf{German-English}} \\
\textbf{Team} & \textbf{BLEU} \\ \hline
MSRA & 42.8 \\
Facebook FAIR & 40.8 \\
NEU & 40.5 \\
\textbf{UCAM} & \textbf{39.7} \\
RWTH &  39.6 \\
MLLP-UPV  & 39.3 \\
DFKI & 38.8 \\
\multicolumn{2}{|c|}{{\em 4 more...}} \\
    \hline
\end{tabular}
\caption{English-German and German-English primary submissions to the WMT19 shared task.}\label{tab:wmt19-leaderboard}
\end{table}

\begin{table}
\centering
\small
\begin{tabular}{|c|c|c|c|}\hline
\textbf{Year} & \textbf{Best in} & \textbf{This work} & \textbf{$\Delta$} \\
 & \textbf{competition} &  &  \\ \hline
2017 & 28.3 & 32.8 & \textbf{+4.5} \\
2018 & 48.3 & 49.3 & \textbf{+1.0} \\
2019 & 44.9 & 43.0 & \textbf{-1.9} \\
    \hline
\end{tabular}
\caption{Comparison of our English-German system with the winning submissions over the past two years.}\label{tab:wmt-sota-ende}
\end{table}

Our WMT19 submission focused on regularized fine-tuning and language modelling. With our novel Intra-Inter Transformer architecture for document-level LMs we achieved significant reductions in perplexity and minor improvements in BLEU over very strong baselines. A combination of checkpoint averaging and EWC proved to be an effective way to regularize fine-tuning. Our systems are competitive on both English-German and German-English (Tab.~\ref{tab:wmt19-leaderboard}), especially considering the immense speed with which our field has been advancing in recent years (Tab.~\ref{tab:wmt-sota-ende}).

\section*{Acknowledgments}

This work was supported by the U.K. Engineering and Physical Sciences Research Council (EPSRC) grant EP/L027623/1 and has been performed using resources provided by the Cambridge Tier-2 system operated by the University of Cambridge Research Computing Service\footnote{\url{http://www.hpc.cam.ac.uk}} funded by EPSRC Tier-2 capital grant EP/P020259/1.

\bibliography{acl2019}
\bibliographystyle{acl_natbib}

%

\end{document}